\documentclass[letterpaper, 10 pt, conference]{ieeeconf}  

\IEEEoverridecommandlockouts                              
\usepackage{graphicx}%
\usepackage{amsmath} 
\usepackage{amssymb}  
\usepackage{tabularx}

\title{\LARGE \bf
 Multi-level Stress Assessment Using Multi-domain Fusion of ECG Signal
}

\author{Zeeshan Ahmad$^{1}$ and	Naimul Mefraz Khan$^{2}$
\thanks{*This work is supported by NSERC}
\thanks{$^{1}$Zeeshan Ahmad is with Department of Electrical and Computer Engineering,
        Ryerson University, 350 Victoria St, Toronto, ON M5B 2K3,Canada.
        {\tt\small z1ahmad@ryerson.ca}}%
\thanks{$^{2}$Naimul Mefraz Khan is with Department of Electrical and Computer Engineering,
	Ryerson University, 350 Victoria St, Toronto, ON M5B 2K3,Canada.
        {\tt\small n77khan@ee.ryerson.ca}}%
}

\begin{document}

\maketitle
\thispagestyle{empty}
\pagestyle{empty}

\begin{abstract}

Stress analysis and assessment of affective states of mind using ECG as a physiological signal is a burning research topic in biomedical signal processing. However, existing literature provides only binary assessment of stress, while multiple levels of assessment may be more beneficial for healthcare applications. Furthermore, in present research, ECG signal for stress analysis is examined independently in spatial domain or in transform domains but the advantage of fusing these domains has not been fully utilized.
To get the maximum advantage of fusing diferent domains, we introduce a dataset with multiple stress levels and then classify these levels using a novel deep learning approach by converting ECG signal into signal images based on R-R peaks without any feature
extraction. Moreover, We
made signal images multimodal and multidomain by converting them
into time-frequency and frequency domain using  Gabor wavelet transform (GWT) and Discrete Fourier Transform (DFT) respectively. Convolutional Neural networks (CNNs) are used to extract features from different modalities and then
decision level fusion is performed for improving
the classification accuracy. The experimental results on an in-house dataset collected with 15 users show that with proposed fusion framework and using ECG signal to image conversion, we
reach an average accuracy of 85.45\%.
\end{abstract}

\section{INTRODUCTION}
ECG is non-invasive physiological signal that is easily available even in commercial wearables such as Apple watch. Studies in~\cite{schmidt2018introducing} and~\cite{giannakakis2019review} show that ECG is a useful signal to measure stress levels. The problem with the existing work is that in most of the works binary classification is performed to measure stress level. For instance, in the popular WESAD dataset~\cite{schmidt2018introducing},
when it comes to stress classification, the data is only grouped into two labels, stress and no-stress.\let\thefootnote\relax\footnote{© 2020 IEEE. Personal use of this material is permitted. Permission from IEEE must be obtained for all other uses, in any current or future media, including reprinting/republishing this material for advertising or promotional purposes, creating new collective works, for resale or redistribution to servers or lists, or reuse of any copyrighted component of this work in other works.}

Furthermore, most of the current studies utilize the
classical supervised learning approach of feature extraction
from heart-rate variablity (HRV), followed by classification.
The rise of deep learning poses an interesting research
question: can deep learning result in improved detection
of stress from ECG? A recent study~\cite{he2019real} reports improved
stress assessment utilizing Convolutional Neural Networks
(CNN). However, this study also utilizes a binary stress
classification approach.

Another shortcoming of the existing works is that ECG signal for stress analysis is utilized independently in spatial domain or in transform domains~\cite{reiss2019deep}, but the benefit of fusing these domains has not been realized. Our recent work in the action recognition domain shows that  fusing multidomain modalities yield better results~\cite{ahmad2019multidomain}.

To address these shortcomings, in this paper, we introduce a dataset with multiple stress levels and in order to classify these levels, we introduce a novel deep learning approach by converting ECG signal into signal images based on R-R peaks without any feature
extraction. In addition to this, we
made signal images multimodal and multidomain by converting them
into time-frequency and frequency domain using  Gabor wavelet transform (GWT) and Discrete Fourier Transform (DFT) respectively. Convolutional Neural networks (CNNs) are used to extract features from different modalities and then
decision level fusion is performed for improving the classification accuracy. Following are our key contributions.

\begin{itemize}
	
	\item We perform a user study with 15 users for multi-level
	stress assessment during a dynamic VR roller coaster
	experience. We collect multiple physiological sensordata, namely, ECG, respiration, and GSR. However, to signify the simplicity and applicability of the proposed approach, in this paper, we only utilize the
	ECG data.

	\item Instead of binary stress classification, we manually
	label the data in 30-seconds intervals from the VR
	roller coaster experience to five levels: very low,
	low, above average, high, very high. Due to the
	varying intensity in the VR experience, manual labeling results in a dataset where there is a rigorous ground
	truth correspondence between ECG and stress levels.

	\item Instead of traditional HRV-based feature extraction
	and classification approach, we propose a novel deep learning approach, where ECG signals are converted to
	images utilizing an image transformation based on
	R-R peaks without any feature extraction, followed
	by CNN-based classification. Conversion to image
	from 1D signals provides us with the opportunity
	to extract higher-level abstract features such as edges
	and blobs, which is not possible in 1D, as we have
	shown previously in the action recognition domain~\cite{ahmad2019human}.
	
\end{itemize}


\section{Related Work}

Deep learning has recently found some success in stress classification.  In~\cite{he2019real}  acute cognitive
stress is classified from ECG signal using CNN. Experiments show that CNN performed better than classical machine learning methods. In~\cite{reiss2019deep}, CNN is used for photoplethysmogram
(PPG) based heart rate estimation. The input to CNN is
the time-frequency spectrum of the PPG-signal and tri-axial
acceleration. In~\cite{uddin2019synthesizing}, a temporal modeling scheme based on
one dimensional CNN and random forest is used to classify,
detect and estimate affective states (stress and meditation)
ahead of time using raw physiological and motion signals.
For discussing the challenges of physiology-based affective
computing in the wild, a multitask CNN is used in~\cite{schmidt2019multi} to
classify arousal, State-Trait Anxiety Inventory (STAI), stress,
and valence self-reports. The results are then compared
with the classical methods using F1 score. In~\cite{chakraborty2019multichannel}, different
biosignals such as blood volume pulse (BVP), wrist-worn acceleration signals, the ECG signal, electromyography (EMG
signal) and respiration are fed to multichannel CNN to
improve the classification and accuracy of different states of
mind. However, specifically for stress assessment, most of
these studies only utilize a binary label (stress/no-stress).
In~\cite{lin2019explainable}, a deep learning based fusion framework is presented to process multimodal-multisensory bio-signals. The
late fusion scheme is used in deep architecture fusion and
model is evaluated by cross validation. The classification
results obtained are used to explain the fact that the chest
modalities contribute more to the correct classification than
the wrist modalities and that the chest EDA has significant
influence on the model. Authors in~\cite{siirtola2019continuous} provided study survey about the possibilities of detecting user-independently
stress accurately using commercial smart watches which
do not include EDA sensor. The effect of window-size on
recognition accuracies of affective states is also investigated
in the article. Performance of different classifiers with different sensor combinations is also tested and reported using the
publicly available WESAD dataset. However, no public dataset exists in literature, which provides us with ground truth labeling of multiple levels of stress, which can be useful for designing dynamic multimedia/healthcare applications. Therefore we collected an in-house dataset.


\section{Data Collection}

Fifteen participants  experienced a 6-minute
rollercoaster simulation in a virtual reality environment. Rollercoasters
have been a means of inducing acute stress. In this
experiment we used a virtual reality simulation to recreate
the real life environment of such an acute stress stimulus.
The participant’s electrocardiogram (ECG), galvanic skin response (GSR) and respiration signals were measured. However, in this paper we only use ECG signal for experiment.
ECG data was collected at sampling frequency of 256 Hz and the the sensor was placed around the
subject’s chest, while keeping the electrodes in contact with
the skin.

The data collected during the roller coaster simulation was
appropriately labeled through observation of the VR experience. Each 30 second segment of the simulation was
assigned a label from 0 to 4 in order of increasing stress
as shown in Table~\ref{tab:groundtruthlabel}.

\begin{table*}[h]
	\caption{Ground Truth Label for Roller Coaster Simulation}
	\label{tab:groundtruthlabel}
	\centering
		
	\begin{tabular}{|c|c|c|}

			\hline 
			\textbf{Label} & \textbf{Label Name} & \textbf{Label Meaning}   \\\hline 
			0  &  Very Low & no stress, nominal conditions \\\hline
			1  &  Low & very slight stress .i.e. the rollercoaster is in dull movement with no loops  \\\hline
			2  &    Above Average & roller coaster makes one or two loops  \\\hline
			3  &   High & roller coaster undergoes multiple events of high activity .i.e. continuous loops, twists, turns and underwater dives  \\\hline
			4  &    Very High & most intense segments .i.e. consisting of surprising jump scares  \\\hline
	\end{tabular}
	
\end{table*}

Such manual labeling solely focuses on the VR experience in hand to label the level of stress, providing a true
correspondence between physiological signals and stress.
The collected dataset will be called the Ryerson Multimedia Research Laboratory (RML) dataset for the rest of the paper.

The experimental procedures involving human subjects described in this paper were approved by the Institutional Review Board (REB \# 2018-462).

\begin{figure}
	\centering
	\includegraphics[width=1\linewidth]{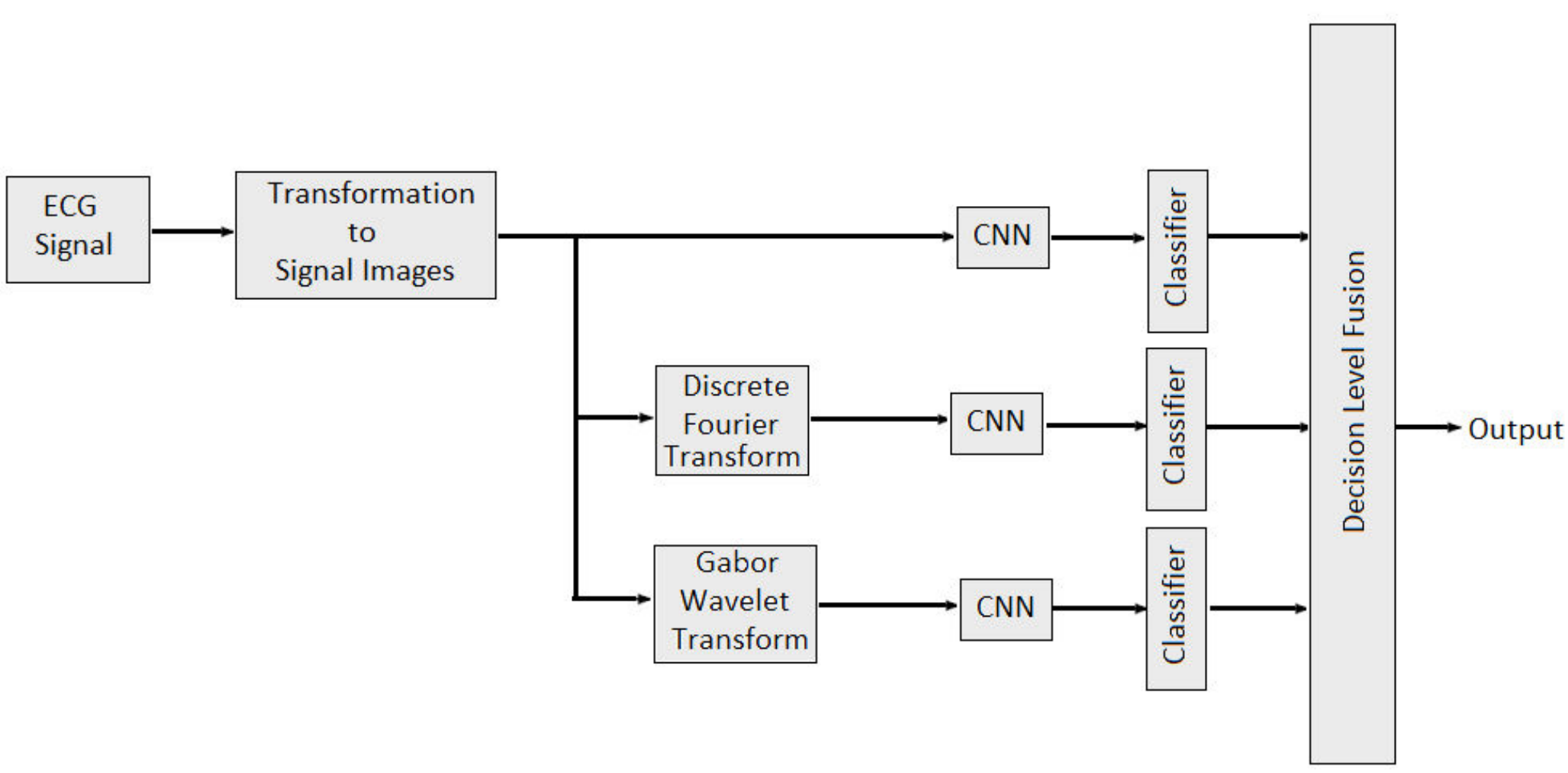}
	\caption{Schematic diagram of Proposed Multi-domain Fusion Framework}
	\label{fig: overview}
\end{figure}

\begin{figure*}
	\centering
	\includegraphics[width=\linewidth]{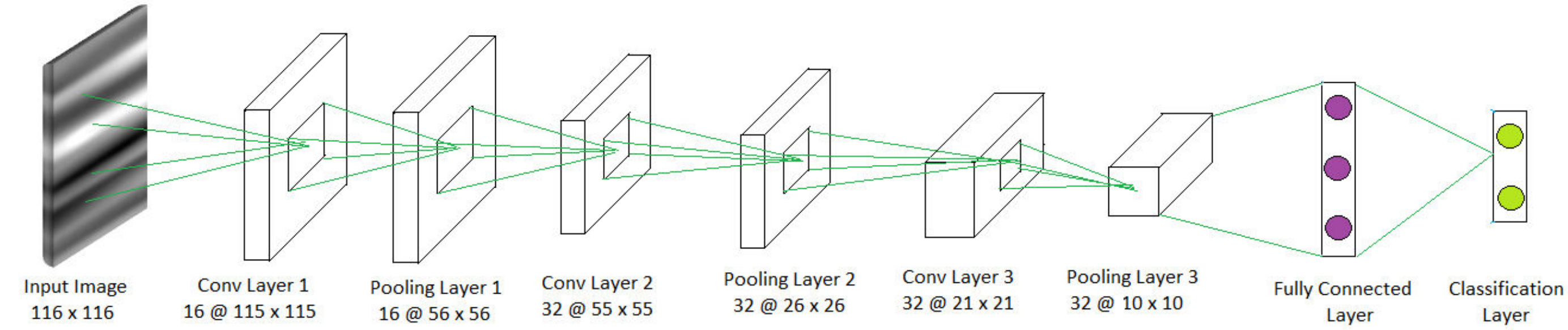}
	\caption{CNN Architecture for Signal Images}
	\label{fig:CNN Architecture for signal images}
	
\end{figure*} 

\begin{figure}[h]
	\centering
	\includegraphics[width=0.9\linewidth]{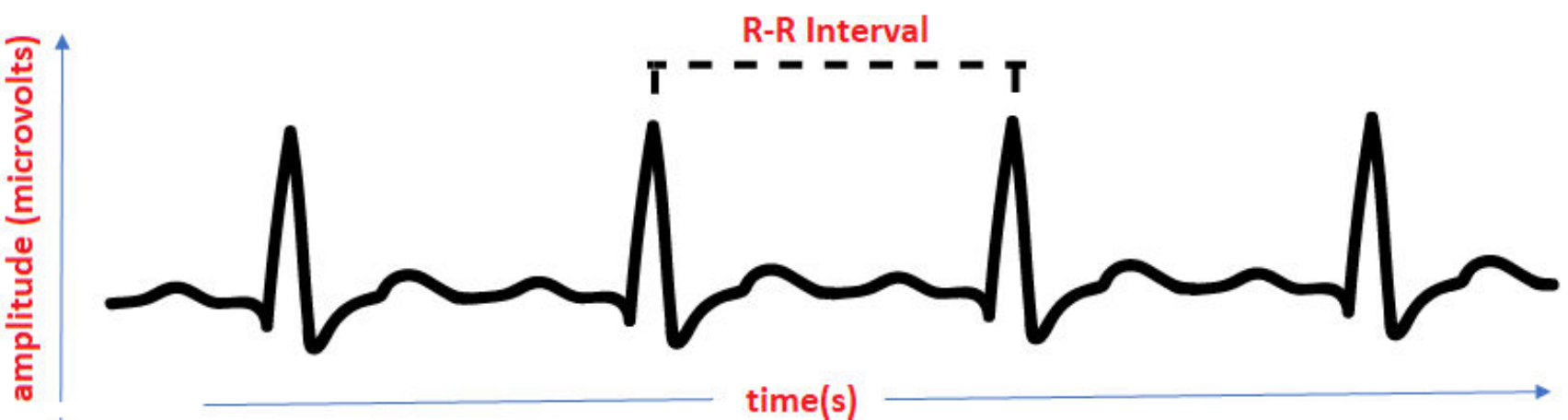}
	\caption{R-R Intervals of ECG signals}
	\label{fig:R-R Intervals}
\end{figure}

\begin{figure}[h]
	\centering
	\includegraphics[width=0.8\linewidth]{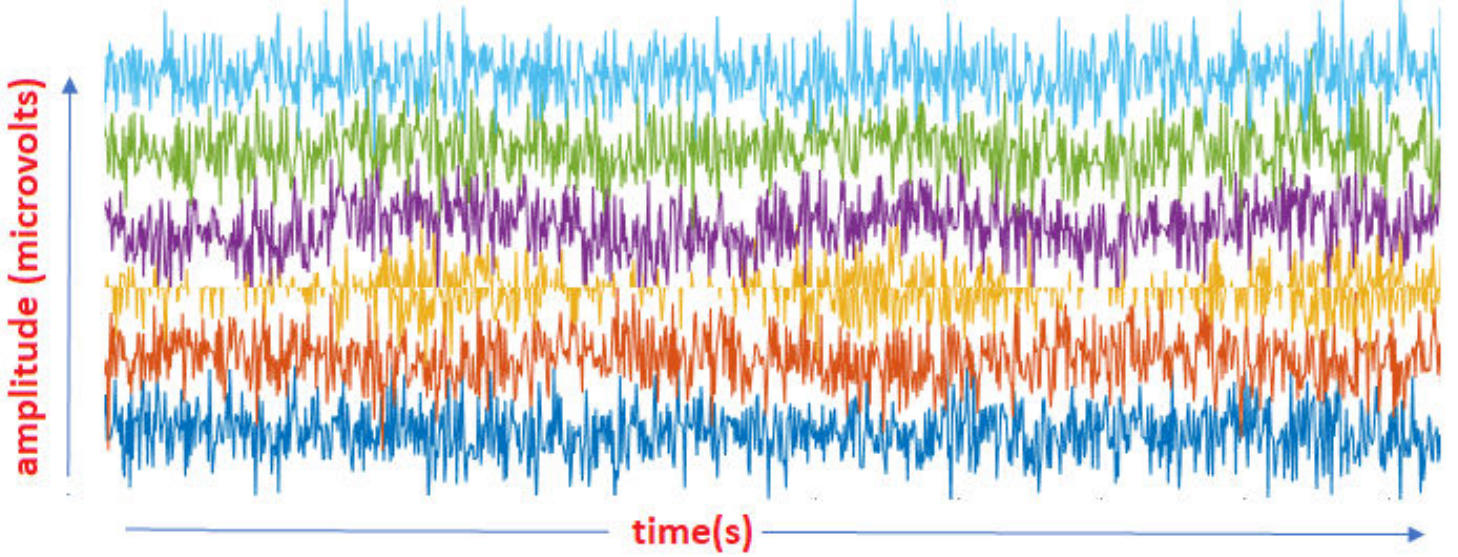}
	\caption{Row by row stacking of ECG segments to form multivariate time series.}
	\label{fig:multivariate ECG Signal}
\end{figure}

\begin{figure}[h]
	\centering
	\includegraphics[width=0.8\linewidth]{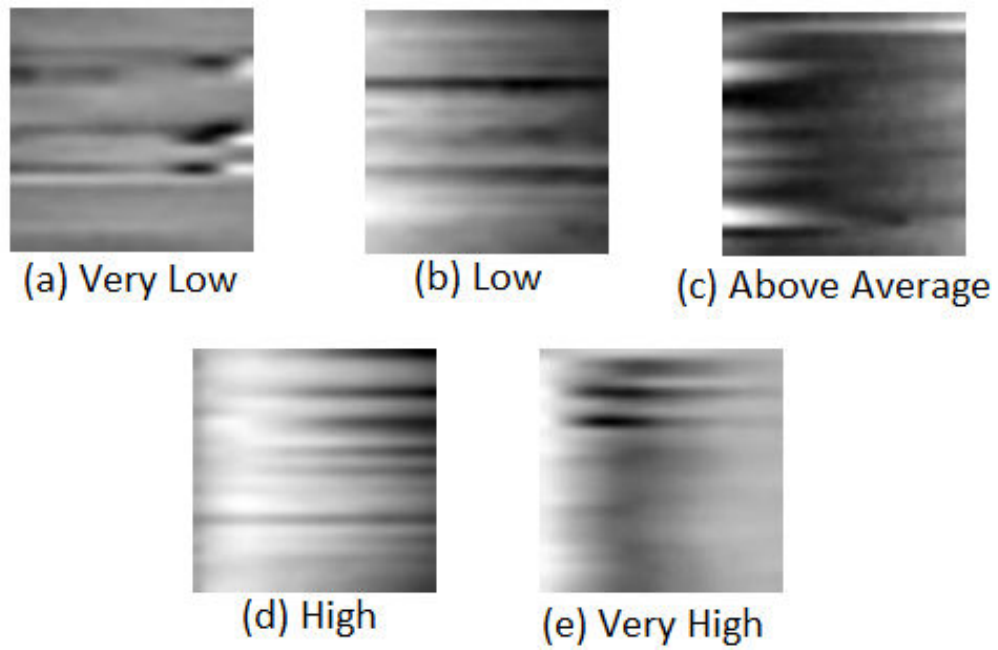}
	\caption{Signal images of Five Stress labels of RML Dataset.}
	\label{fig:Signal Image}
\end{figure}	


\section{Proposed Multi-domain Fusion Framework}\label{proposed method}

At the input of the proposed multi-domain fusion framework, ECG signal is converted into signal images using temporal correlation between each successive R-R interval. Then, these signal images are  made multidomain by converting spatial domain information into frequency and time-frequency domain by applying DFT and GWT using equations~\ref{FFT} and ~\ref{Gabor} respectively. CNNs, whose structure is laid out in Fig.~\ref{fig:CNN Architecture for signal images}, are engaged to pull out multi-domain features from the modalities and then decision level fusion is performed to improve the accuracy of stress levels classification. Schematic diagram of proposed muti-domain fusion framework is shown in Fig.~\ref{fig: overview}.

\subsection{ Discrete Fourier Transform}

2D DFT transforms input spatial domain image $f(m,n)$ into frequency domain image $F(k,l)$. Each point in frequency domain is a representation of a particular frequency contained in the spatial domain image.

\begin{equation} \label{FFT}
F(k,l) = \sum_{m=0}^{a-1}\sum_{n=0}^{b-1}{f(m,n)}e^{-j2\pi(\frac{km}{a}+\frac{ln}{b})}
\end{equation}

Where $f(m,n)$ is the spatial domain image of size $a \times b$.      

\subsection{Gabor Wavelet Transform}

GWT transforms the images into time-frequency domain. 2D Gabor filter used for image conversion is bi-dimensional 
Gaussian function centered at origin (0,0) with variance $\sigma$ modulated by
a complex sinusoid with polar frequency $(F,\omega)$ and phase $\theta$ described in equation~\ref{Gabor}.

\begin{equation} \label{Gabor}
GWT =Ae^{-\pi\sigma^2(x^2+y^2)}e^{(j2\pi F(xcos\omega+ysin\omega)+\theta)}
\end{equation}
In equation~\ref{Gabor},  $A$ represents magnitude of Gaussian envelope.

\begin{figure}[h]
	\centering
	\includegraphics[width=0.7\linewidth]{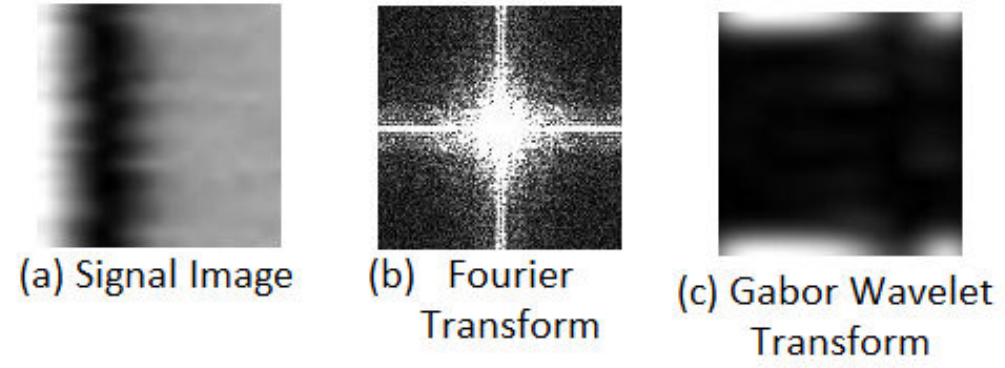}
	\caption{Signal Image and its Fourier and Gabor Wavelet Transform. }
	\label{fig:transform of images.}
\end{figure}

The signal image and its Fourier and Gabor wavelet transform are shown in Fig.~\ref{fig:transform of images.}

\subsection{Signal Image Formation from ECG}\label{Signal Image formation}

The ECG signal obtained from chest-worn sensor is a univariate time series consists of large number of R-R inervals as shown in Fig.~\ref{fig:R-R Intervals}. In order to convert univariate time series into multivariate form, we make segments of time series between each successive R-R intervals and stacked these segments row wise as shown in Fig.~\ref{fig:multivariate ECG Signal}.

Since the number of signal samples between successive R-R interval are not the same, we perform interpolation to make the number of samples equal between successive R-R interval and hence to make row wise stacking possible. While Interpolating,
we take care of the fact that R-R interval and HRV are
preserved.

Multivariate time series obtained by row wise stacking of segments between R-R intervals has a strong temporal correlation among its rows. Signal images are formed by taking advantage of temporal correlation among these stacked segments as shown in Fig.~\ref{fig:Signal Image}. These signal images are 2D representations carrying more semantic information than original signal. Furthermore, converting 1D time series data to 2D representation enables CNN to pull-out more distinctive low level and high level features that are not possible with ID signal.

\subsection{Multidomain Fusion}\label{multidomain fusion}

CNNs are employed to extract multidomain features and perform stress labels classification. Finally, we perform decision level fusion by using the scores generated by classifiers to improve the classification performance as shown in Fig.~\ref{fig: overview}.

\begin{table}[h]
	\caption{Training Parameters for CNN}
	\label{tab:parameters for CNN}
	\centering
	\begin{tabular}{|c|c|}
		
		\hline 
		\textbf{Training Parameters} & \textbf{Values}   \\\hline 
		Momentum  &      0.9 \\\hline
		Initial Learn Rate  &      0.005 \\\hline
		Learn Rate Drop Factor  &      0.5 \\\hline
		Learn Rate Drop Period  &      10 \\\hline
		$L_2$ Regularization  &      0.004 \\\hline
		MiniBatchSize  &   64 \\\hline				
	\end{tabular}
	
\end{table}

\begin{table}[h]
\caption{Results of Experiments on RML Dataset}
\label{experiments on RML Dataset}	
	\centering
		\begin{tabular}{|c|c|}	
			\hline 
			\textbf{Experiments} & \textbf{Accuracy\%}   \\\hline
			Raw ECG Signal with 1D CNN  &    64.8 \\\hline 
			Signal Images only  &      83.6 \\\hline
			FFT Images only & 74.5   \\\hline
			Gabor Images only  &  78.1     \\\hline
			Decision Level Fusion & 85.45 \\\hline 
			
	\end{tabular}
	
\end{table} 


\section{Experimental Results}

For experiments, we
divide the dataset into training and testing parts by randomly
splitting 85\% data into training samples and 15\% data into testing
samples. To facilitate the design of CNN shown in figure~\ref{fig:CNN Architecture for signal images}, we resize our images to 116 x 116. Size of training samples is increased by performing data augmentation on signal images using the same augmentation techniques described in~\cite{ahmad2018towards}.
Thus 1350 images are used for training while 55 images are used for testing. We execute the random split 10 times
and report the average accuracy. We also perform a baseline experiment on raw ECG signal using 1D CNN. The input is a vector of size 116 x 1. The accuracy obtained is shown in Table~\ref{experiments on RML Dataset}.
The poor performance of 1D CNN on raw ECG data compelled us to transform the raw 1D data into more informative i.e 2D form. Thus to improve baseline experiments,
we transform raw ECG signal into signal images and utilize CNNs to extract features as CNN is popular deep learning network to perform exceptional on computer vision and image classification applications.
We train CNN on the
signal images till the validation loss stops decreasing further. The values of other training parameters for CNN are shown in Table~\ref{tab:parameters for CNN}.
Table~\ref{experiments on RML Dataset} shows that convering the ECG signal to images results in a higher accuracy compared to the baseline 1D CNN. The accuracy obtained in transform domains is less than spatial domain. The reason for this reduced accuracy is that during transforms, we lose information that varies with time. Furthermore in ECG signal, information or features varies abruptly near R-R intervals and that abrupt changes cannot be modeled accurately in transform domains. However, these domains provide useful complementary features and hence the overall accuracy is improved by decision level fusion.


\section{Conclusion}

In this paper, we propose a novel deep learning approach by converting input ECG signal into images based on R-R peaks and temporal correlation. We transform spatial domain signal images into transform domains by applying Discrete Fourier Transform and Gabor wavelet transform respectively and extract unique, complementary and multidomain features using CNNs. Finally we perform decision level fusion to enhance the accuracy of stress levels classification. We also introduce a dataset with multiple stress levels that can be more useful than the existing binary levels datasets for stress analysis. Experimental results on an in-house dataset with ground truth label of 5 different stess levels show that our proposed fusion method achieves the highest accuracy of 85.45\% compared with a baseline 1D CNN, and the separate image modalities. 

\addtolength{\textheight}{-12cm}

\section{ACKNOWLEDGMENT}

NSERC’s financial support for this research work through
a Collaborative Research and Development (CRD) Grant (\#
537987-18) is much appreciated.



\begin{thebibliography}{99}

\bibitem{schmidt2018introducing}
P.~Schmidt, A.~Reiss, R.~Duerichen, C.~Marberger, and K.~Van~Laerhoven,
``Introducing wesad, a multimodal dataset for wearable stress and affect
detection,'' in \emph{Proceedings of the 2018 on International Conference on
	Multimodal Interaction}.\hskip 1em plus 0.5em minus 0.4em\relax ACM, 2018,
pp. 400--408.

\bibitem{giannakakis2019review}
G.~Giannakakis, D.~Grigoriadis, K.~Giannakaki, O.~Simantiraki, A.~Roniotis, and
M.~Tsiknakis, ``Review on psychological stress detection using biosignals,''
\emph{IEEE Transactions on Affective Computing}, 2019.

\bibitem{he2019real}
J.~He, K.~Li, X.~Liao, P.~Zhang, and N.~Jiang, ``Real-time detection of acute
cognitive stress using a convolutional neural network from
electrocardiographic signal,'' \emph{IEEE Access}, vol.~7, pp.
42\,710--42\,717, 2019.

\bibitem{reiss2019deep}
A.~Reiss, I.~Indlekofer, P.~Schmidt, and K.~Van~Laerhoven, ``Deep ppg:
Large-scale heart rate estimation with convolutional neural networks,''
\emph{Sensors}, vol.~19, no.~14, p. 3079, 2019.

\bibitem{ahmad2019multidomain}
Z.~Ahmad and N.~M. Khan, ``Multidomain multimodal fusion for human action
recognition using inertial sensors,'' in \emph{2019 IEEE Fifth International
	Conference on Multimedia Big Data (BigMM)}.\hskip 1em plus 0.5em minus
0.4em\relax IEEE, 2019, pp. 429--434.

\bibitem{ahmad2019human}
Z.~Ahmad and N.~Khan, ``Human action recognition using deep multilevel
multimodal (m2) fusion of depth and inertial sensors,'' \emph{IEEE Sensors
	Journal}, 2019.

\bibitem{uddin2019synthesizing}
M.~T. Uddin and S.~Canavan, ``Synthesizing physiological and motion data for
stress and meditation detection,'' in \emph{2019 8th International Conference
	on Affective Computing and Intelligent Interaction Workshops and Demos
	(ACIIW)}.\hskip 1em plus 0.5em minus 0.4em\relax IEEE, 2019, pp. 244--247.

\bibitem{schmidt2019multi}
P.~Schmidt, R.~D{\"u}richen, A.~Reiss, K.~Van~Laerhoven, and T.~Pl{\"o}tz,
``Multi-target affect detection in the wild: an exploratory study,'' in
\emph{Proceedings of the 23rd International Symposium on Wearable Computers},
2019, pp. 211--219.

\bibitem{chakraborty2019multichannel}
S.~Chakraborty, S.~Aich, M.-i. Joo, M.~Sain, and H.-C. Kim, ``A multichannel
convolutional neural network architecture for the detection of the state of
mind using physiological signals from wearable devices,'' \emph{Journal of
	healthcare engineering}, vol. 2019, 2019.

\bibitem{lin2019explainable}
J.~Lin, S.~Pan, C.~S. Lee, and S.~Oviatt, ``An explainable deep fusion network
for affect recognition using physiological signals,'' in \emph{Proceedings of
	the 28th ACM International Conference on Information and Knowledge
	Management}, 2019, pp. 2069--2072.

\bibitem{siirtola2019continuous}
P.~Siirtola, ``Continuous stress detection using the sensors of commercial
smartwatch,'' in \emph{Adjunct Proceedings of the 2019 ACM International
	Joint Conference on Pervasive and Ubiquitous Computing and Proceedings of the
	2019 ACM International Symposium on Wearable Computers}, 2019, pp.
1198--1201.

\bibitem{ahmad2018towards}
Z.~Ahmad and N.~Khan, ``Towards improved human action recognition using
convolutional neural networks and multimodal fusion of depth and inertial
sensor data,'' in \emph{2018 IEEE International Symposium on Multimedia
	(ISM)}.\hskip 1em plus 0.5em minus 0.4em\relax IEEE, 2018, pp. 223--230.

\end{thebibliography}
\end{document}